\documentclass{article}

\usepackage{arxiv}

\usepackage[utf8]{inputenc} % allow utf-8 input
\usepackage[T1]{fontenc}    % use 8-bit T1 fonts
\usepackage{hyperref}       % hyperlinks
\usepackage{url}            % simple URL typesetting
\usepackage{booktabs}       % professional-quality tables
\usepackage{amsfonts}       % blackboard math symbols
\usepackage{nicefrac}       % compact symbols for 1/2, etc.
\usepackage{microtype}      % microtypography
\usepackage{lipsum}		% Can be removed after putting your text content
\usepackage{graphicx}
\usepackage{natbib}
\usepackage{multirow}
\usepackage{doi}

\title{Can We Automate the Analysis of Online Child Sexual Exploitation Discourse?}

\date{}	% Here you can change the date presented in the paper title
\author{Darren Cook \\
  University of Liverpool \\ Liverpool, UK \\
  \texttt{d.cook@liverpool.ac.uk} \\\And
  Miri Zilka \\
  University of Cambridge \\ Cambridge, UK \\
  \texttt{mz477@cam.ac.uk} \\\And
  Heidi DeSandre \\
  University of Liverpool \\ Liverpool, UK \\
\And
  Susan Giles \\
  University of Liverpool \\ Liverpool, UK \\
  \And
  Adrian Weller \\
  University of Cambridge \\ The Alan Turing Institute \\ UK \\
  \And
  Simon Maskell \\
  University of Liverpool \\ Liverpool, UK \\
  }

\renewcommand{\shorttitle}{Can We Automate the Analysis of Online Child Sexual Exploitation Discourse?}

\begin{document}
\maketitle

\begin{abstract}
Social media's growing popularity raises concerns around children's online safety. Interactions between minors and adults with predatory intentions is a particularly grave concern. Research into online sexual grooming has often relied on domain-experts to manually annotate conversations, limiting both scale and scope. In this work, we test how well automated methods can detect conversational behaviors and replace an expert human annotator. Informed by psychological theories of online grooming, we label $6772$ chat messages sent by child-sex offenders with one of eleven predatory behaviors. We train bag-of-words and natural language inference models to classify each behavior, and show that the best performing models classify behaviors in a manner that is consistent, but not on-par, with human annotation.
\end{abstract}

% keywords can be removed
% \keywords{First keyword \and Second keyword \and More}

\section{Introduction}
Social media's growing popularity amongst children and young adults raises serious concerns for their safety. The threat of online sexual grooming is an increasing problem in the digital age \citep{greene2020experiences}. In 2021 alone, UK police forces recorded over 5000 offences relating to sexual communications with a child, with social media apps such as Instagram and Snapchat popular amongst online predators \citep{nspcc}. Minors who fall victim to offenders suffer considerably, with many abusers seeking to establish physical contact offline \citep{shelton2016online}. To assist law enforcement, academics have been trying to identify in advance those predators that steer the relationship towards physical encounters \citep{briggs2011exploratory, winters2017sexual, o2003typology, williams2013identifying}. This important work predominantly relies on time-consuming human-effort to manually annotate conversations involving child sex offenders. In this work, we explore the extent to which automated annotation can mimic an expert human annotator.

% Automatic detecting online behaviors is a challenging task for several reasons. Offenders engage in a variety of behaviors when grooming children, ranging from flattery to threats \cite{joleby2021offender}. This means complex or subtle behaviors may not be captured by simplistic methods (i.e., counting all the sex-based words). Detecting instances of grooming would be useful for law enforcement. However, a majority of studies focus on detecting predatory chats after-the-fact \cite{razi2021human} and are thus of limited utility in a real-world setting.

Automatically detecting predatory behavior is a challenging task. Offenders use a variety of subtle behaviors to manipulate the flow of the conversation. Predators may use flattery to build trust \citep{barber2021exposing}, make threats or bribe a child as a coercion tactic \citep{joleby2021offender}. Whilst human experts can identify these contextual psychological behaviors in text, their implicitness can be problematic for machines \citep{buckingham2020extreme}. In computational social sciences, dictionary-based approaches \citep{tausczik2010psychological} are often used to identify psychological characteristics such as neuroticism \citep{bogdanova2014exploring}. However, these methods heavily rely on the included vocabulary, causing a large numbers of false positives for some behaviors \citep{kaur2021abusive} while overlooking others \citep{broome2020psycho}. Prior work utilising machine learning has focused on identifying predators from a mixed corpus of illicit and everyday conversations \citep{inches2012overview,pendar2007toward,miah2011detection,EBRAHIMI201633,gupta2012characterizing}. While valuable in its own right, this line of research does not offer significant value to grooming experts as it identifies grooming behavior after-the-fact \citep{razi2021human} and does not link or contribute to psychological insight.  

% Software such as the Linguistic Inquiry Word Count (LIWC: \cite{tausczik2010psychological}) has sought to overcome this by generating word lists for a range of psychological behaviors. LIWC has become a popular tool within the computational social sciences, where it has been used to identify characteristics such as neuroticism \cite{bogdanova2014exploring}. However, a criticism of dictionary-based approaches is an over-reliance on the included vocabulary. This can cause large numbers of false positives for some categories \cite{kaur2021abusive}, while domain-specific behaviors can be overlooked \cite{broome2020psycho}.

% Machine learning research has often concentrated on identifying predators from a mixed corpus of illicit and everyday conversations \cite{inches2012overview}. \citet{pendar2007toward} used a k-Nearest Neighbour approach to identify predators from victims of child grooming. \citet{miah2011detection} used term frequencies and supervised ML to classify predatory chats, while other approaches include using Convolutional Neural Networks (CNN) \cite{EBRAHIMI201633}, vector space models \cite{fauzi2020ensemble}, and network graphs \cite{gupta2012characterizing}. Whilst performance metrics for these studies are often encouraging, their approaches necessitate detection of grooming behavior after-the-fact \cite{razi2021human}. As such they are of limited utility in a preventative setting. A reliance on atheoretical linguistic features also limits the interpretability of the predictions made.

This work is novel in its use of supervised and deep learning methods to analyse an expert-annotated corpus of real conversations between sexual predators and decoys pretending to be early teens. We report results and highlight the performance possible without incurring the cost of expert-labelling.
The paper is organised as follows: the methods and experimental setup are described in Section~\ref{sec:method}; Section~\ref{sec:results} then summarises results (with further details provided in the appendices) before Section~\ref{sec:concl} concludes the paper.
%  We report the models' performance in identifying eleven communication techniques used by the offender on a hold-out set. Considering the significant cost of expert-labeling, we investigate the improvement in performance when going from zero to few-shot settings \cite{wang2021entailment, yin2020tryfewshot}. We show that the best performing model achieves an inter-annotator agreement $\sim 20\%$ lower than the agreement between two human annotators. 

\section{Method and Experimental Setup}
\label{sec:method}

\begin{table}[h]
\centering
\begin{tabular}{lcc}
\hline
\textbf{Region} & \textbf{Num. Msgs} & \textbf{\% of Corpus}\\
\hline
Train & 4712 & 70\% \\
Test & 1355 & 20\% \\
Validation & 704 & 10\%\\
\hline
\end{tabular}
\caption{The number of messages included for training, testing, and validation regions.}
\label{tab:data-splits}
\end{table}

\begin{table*}[h]
\centering
\small{
\begin{tabular}{lccccc}
\hline
\textbf{Category} & \textbf{Coverage (\%)}& \textbf{Model}& \textbf{Precision}& \textbf{Recall} & \textbf{F1} \\
\hline
Communication/Coordination & 73.1 & NLI (1) & $84.0$ ($\pm0.6$) & $87.1$ ($\pm0.5$) & $85.5$ ($\pm0.2$)\\
Rapport Building & 15.2 & NLI (5) & $83.0$ ($\pm9.3$)  & $80.1$ ($\pm15.7$)  & $81.3$ ($\pm12.6$) \\
Control & 20.8 & NLI (5) & $61.5$ ($\pm1.7$) & $59.5$ ($\pm1.4$) & $60.4$ ($\pm0.2$)\\
Challenge & 4.5 & NLI (5) & $36.1$ ($\pm1.7$) & $24.2$ ($\pm4.1$) & $28.9$ ($\pm3.5$)\\
Negotiation & 20.9 & NLI (5) & $62.9$ ($\pm2.9$) & $66.7$ ($\pm1.7$) & $64.7$ ($\pm2.3$)\\
Use of Emotions & 16.4 & NLI (5) & $58.2$ ($\pm1.4$) & $53.1$ ($\pm1.65$) & $55.6$ ($\pm1.6$)\\
Testing Boundaries & 31.2 & NLI (5) & $69.6$ ($\pm0.8$) & $76.1$ ($\pm2.7$) & $72.7$ ($\pm0.9$)\\
Use of Sexual Topics & 18.3 & NLI (5) & $64.5$ ($\pm1.6$) & $69.3$ ($\pm3.6$) & $66.8$ ($\pm2.1$)\\
Mitigation & 3.0 & NLI (3) &   $62.2$ ($\pm10.7$) & $38.4$ ($\pm7.0$) & $47.5$ ($\pm8.5$)\\
Encouragement & 8.0 & NLI (1) & $43.2$ ($\pm4.8$) & $22.0$ ($\pm1.6$) & $29.1$ ($\pm2.0$)\\
Risk Management & 4.6 & NLI (1) & $63.1$ ($\pm4.0$) & $49.5$ ($\pm4.1$) & $55.4$ ($\pm3.5$)\\
\hline
\end{tabular}}
\caption{Eleven offender behavior categories derived and labelled by domain-experts, alongside their prevalence within the corpus. We report performance for the best performing model (based on F1) for each category. NLI (5) denotes the 5-message-input Natural Language Inference model, and similarly for NLI (3) and NLI (1). }
\label{tab:main-results}
\end{table*}

% \subsection{Labelling Framework}
% Chat transcripts were labelled by domain-experts with a background in forensic psychology\footnote{Labelling was performed by the third and fourth author.}. 

\paragraph{Dataset and Labelling.}
The \citet{pjwebsite} website  is an online repository of real, chat-based conversations between adults who were later convicted of grooming offences and decoys posing as children. Twenty-four chats, comprising of 12,942 messages, were labelled by a domain-expert with a background in forensic psychology. Chats were annotated in concordance with a theory of child grooming known as ``self-regulation'' \citep{elliott2017self} -- the notion that online predation contains a \textit{potentiality phase}, where the predator attempts to form a positive relationship with the victim; and a \textit{disclosure phase}, where the predator becomes more explicitly goal-oriented. 
%Through a grounded theoretical approach, 
The following behaviors were identified and applied to the offender messages: (1) communication/coordination, (2) rapport building, (3) control, (4) challenges, (5) negotiation, (6) use of emotions, (7) testing boundaries, (8) use of sexual topics, (9) mitigation, (10) encouragement, (11) risk management. A qualitative description of these behaviors is given in Section~\ref{app-label-desc} of the Appendix. Overall 6,772 messages sent by the offender were labelled, with each message assigned a `Yes' or `No' for each of the above categories, based on the annotator's judgment. Due to the subjective nature of this assessment, labelling performed by different expert annotators may not be in complete agreement. 

We construct binary classification tasks and predict whether each message is an example of each behavior category. Offender messages were split into training, testing, and validation regions (see Table~\ref{tab:data-splits} for information on data splits), and were stratified to ensure equal distribution of behaviors per region. To increase model confidence, we cross validated each experiment three times through random re-sampling. In our results, we report the average ($\pm SD$) scores for precision, recall and F1 as evaluation metrics. 
% Data processing and experimental code was written in python and will be made publicly available on GitHub.

% \subsection{Keywords}\label{keywords}
% Dictionary of related terms generated by domain-experts and validated on a different sample of CSE chats. Each message was one-hot encoded. See Appendix~\ref{keyword-app} for a full list of keywords.

\paragraph{Supervised Machine Learning.} Offender messages were tokenized, part-of-speech (POS) tagged, and lemmatized with \texttt{spaCy}. We removed stop-words (i.e., the, a, am, at, be, is), and used the frequency counts of the remaining unigrams as input features. Consistent with previous work \citep{miah2011detection, bogdanova2014exploring}, we compared the performance of four classifiers: Random Forest, Logistic Regression, Support Vector Machines, and Naive Bayes. For each algorithm, we optimized hyperparameters\footnote{Hyperparameters and the range of values explored are included in Section~\ref{hyp-app} of the Appendix.} using the GridSearch class in \texttt{Scikit-learn} \citep{pedregosa2011scikit} with 3-fold cross validation.

\paragraph{Deep Learning.}
To make predictions using transformer-based deep learning, we formulate the problem as a natural language inference (NLI) task. In NLI, the objective is to determine whether two sentences logically complement or contradict one another \citep{bowman2015large}. For our purposes, we compare each message with a standardized sentence crafted from each behavior label (i.e., ``This message is an example of \textit{building rapport}''). Each message/label sentence pair is then used as an input to the deep learning model, where a softmax activation function calculates the probability the two sentences are logically related. For training\footnote{We used a Tesla P100 GPU for training.}, we use \texttt{RoBERTa-large} \citep{liu2019roberta} with an implementation built in Pytorch. This model has been fine-tuned for NLI tasks using the Multi-Genre Natural Language Inference corpus \citep{N18-1101}. We ran both zero-shot and few-shot training setups. For the zero-shot model, predictions were made on the test set without any domain-specific training. In few-shot settings, we experiment with different amounts of positive training examples up to the full training set. As per \citep{wang2021entailment}, models were trained for 10 epochs with batch size 32 and a learning rate of $10^{-5}$.

\paragraph{Multi-Message Input.}
To explore whether the surrounding messages increase the contextual understanding of the transformer, we repeat full-shot experiments but expand the message window to include multiple prior messages sent by both speakers. To do this, we concatenate multiple messages into a single input which is then passed to the transformer. We compare the performance of NLI models with an input of 1, 3, and 5 messages.

\paragraph{Human-in-the-Loop Validation.}
% In \citet{razi2021human}, it is claimed that relying on standard ML metrics alone, while useful for evaluating algorithmic performance, is insufficient with regards to end-users' satisfaction with the predictions made.

To establish how comparable the model's output is to labelling performed by a human annotator, we scored the predictions made by the best performing model for $\sim10\%$ of the corpus ($645$ predator messages). Each message was scored on a 1-3 scale of agreement (1=disagree, 2=uncertain, 3=agree). Sections of chats were chosen at random. To mitigate potential bias, annotations generated during the initial labelling were shielded from view.

We use Cohen's $\kappa$ \citep{cohen1960coefficient} to measure pairwise agreement between the two human raters and the predictions generated by the best performing model for each behavior. We report the mean of these scores as an index of overall agreement.

\section{Results}
\label{sec:results}

Transfomer-based deep learning models achieved best performance across all categories, with the 5 message input NLI model performing best for most categories. Evaluation metrics for the highest performing model for each behavior are shown in Table~\ref{tab:main-results}. The performance of all models on all categories is reported in Section~\ref{full-results} of the Appendix. Eight of the eleven behaviors obtain a maximum F1 score above 50\%, with the most prevalent category, \textit{communication/coordination}, performing best with F1=$86\%$. Automatically labelling \textit{encouragement} and \textit{challenge} was least successful, with a maximum F1 score of $29\%$ for both. 
For all eleven categories, a transformer-based model, trained on the full training set, was more successful than the traditional supervised machine learning algorithms. Averaging across behaviors, transformers improved performance of single message classification from $33\%$ to $53\%$ compared to the best performing supervised model. In 8 out of 11 cases, the supervised models, trained on the full training set, outperformed the zero-shot transformer. However, in 10 out of 11 cases, only 50 labelled instances were required by the transformer to exceed the best performing bag-of-words models. As many of the behaviors rely on contextual understanding, we increased the input size from a single message to 3 and 5 messages. The first contains two messages sent by the offender and one by the decoy, and the latter contains three by the offender and two by the decoy. Increasing the window size indeed improved performance for all behaviors\footnote{The majority of messages ($73\%$) were labelled as \textit{communication/coordination}. As a result, this behavior was not tested in a multi-message setting as grouping messages meant that all inputs would be positively labelled.} except for \textit{encouragement} and \textit{risk management}. On average, while 3-message classification did not improve single-message performance, 5-message classification improved on single-message F1 from $50\%$ to $56\%$.
% categories that rely on a back and forth seemed to be improved by the larger window -i.e., rapport-building increased F1 by over 20\% between single and 5 msg  --- testing boundaries increased by up to 9\%.

\begin{figure}[h]
\centerline{\includegraphics[scale=0.55]{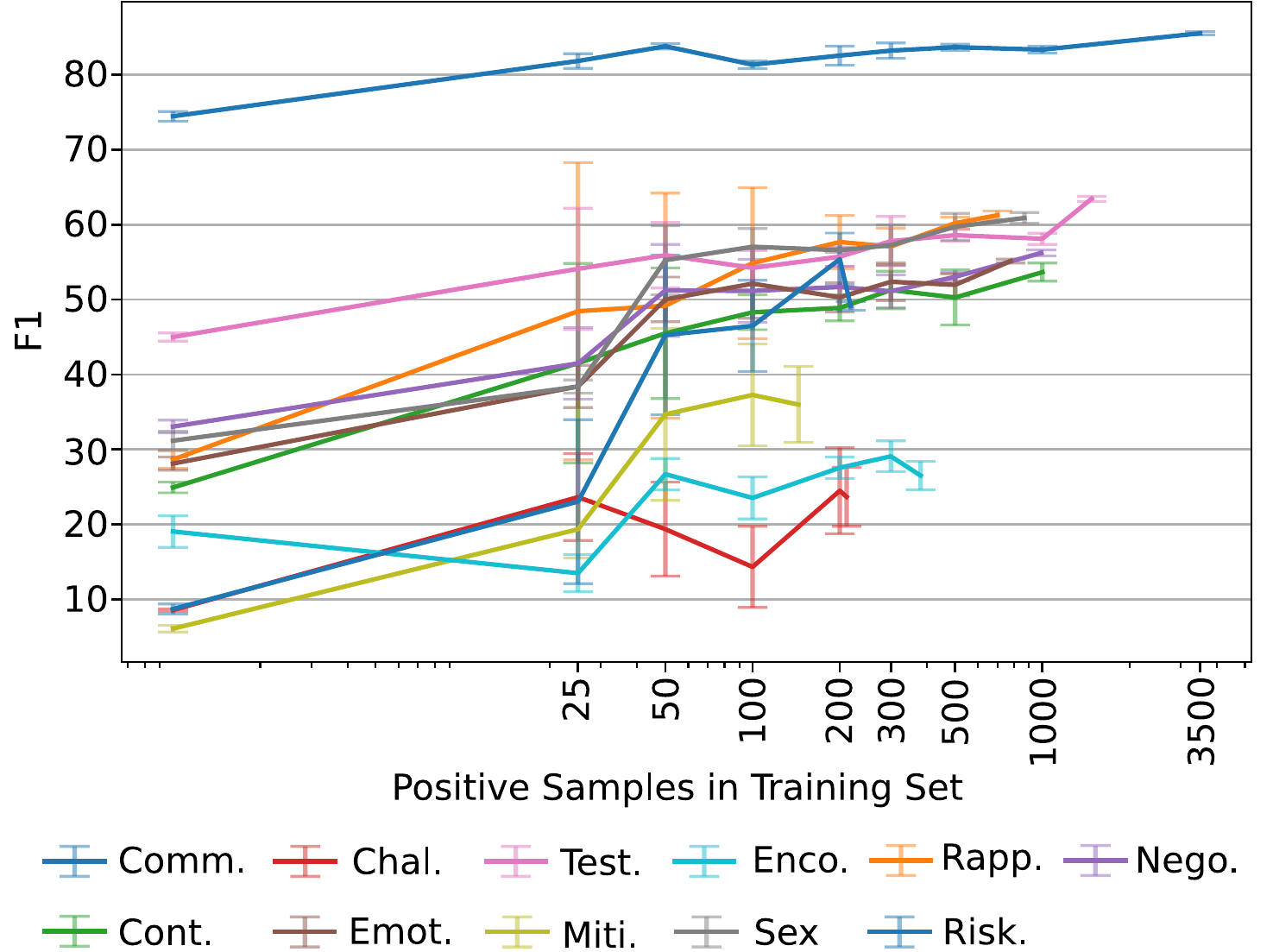}}
\caption{Model performance (F1) on the eleven behavioral categories for single message classification, as the number of positive training examples increases (up to full training set). Error bars are standard  deviations.}
\label{fewshot-perf-change}
\end{figure}

% This may due to inherent difficulty, or an inadequate number of training examples (380 and 211 positive examples respectively). 
Figure~\ref{fewshot-perf-change} shows the change in F1 for single message classification as the number of training samples increases. Whilst predictions for all behaviors improved with additional training, large variation in performance between the categories remains even for a fixed amount of positive training examples. 
To better understand this variation, we manually examined the predictions made by the best performing single message model for $\sim 10\%$ of the corpus. We observed that the transformer has learned some rules that caused a number of false positives. For example, the \textit{testing boundaries} model correctly predicted a number of questions aimed at assessing the victim's willingness to engage, e.g., "what are you looking for?", but also included several logistical questions, e.g., "how about i come by at 8pm?". The \textit{rapport building} model correctly recognised complements and sweet talk as positive examples, but missed more everyday examples of rapport building, e.g., "how was your spring break?". Some aspects of \textit{control} required coverage over longer ranges than we have focused on. For example, persistently asking the same question is a sub-category of \textit{control} that was often misclassified. In this case, increasing the the message window to 3 messages was not enough to improve performance, however, the larger 5 message window did notably improve performance, with the F1 increasing from $54\%$ to $60\%$ due to better recall. In trying to predict \textit{encouragement}, the worst performing category, the transformer appeared to overfit on short verbal nods (i.e., "kool" and "sure", which appeared in the corpus either as evidence of encouragement or simple linguistic fillers) which increased the false positive rate. Over-reliance on this simple rule would explain why the model did not utilise the additional information provided in the multi-message setting. \textit{Risk management} achieved reasonable performance (F1=$55\%$) despite being one of the rarest behaviors in the corpus. This was largely a consequence of the transformer recognising instances where the offender attempts to establish the presence of a parent, e.g., ``when are they getting home?''.

In addition to drawing qualitative observations on the ability of the transformer-based model to recognise different offender behaviors, we quantify how comparable the best performing single-message model is to a human annotator. We compare the labelled classes between the original labels (Rater 1), with the predictions made by the best performing models (Model), and the labels verified during post-validation (Rater 2). We report a mean $\kappa$ score of 0.68, indicating `substantial agreement' between raters \citep{mchugh2012interrater}. Pairwise agreement is reported in Table~\ref{tab:post-val}. Agreement was highest between the two human raters ($\kappa=0.8$). The average agreement between each rater and the predictions generated by the model was $\kappa=0.62$ (see Section~\ref{app:agree} in the Appendix for pairwise agreement per behavior).

\begin{table}[h]
\centering
\begin{tabular}{lccc}
\hline
 & \textbf{Rater 1} & \textbf{Rater 2} & \textbf{Model}\\
\hline
\textbf{Rater 1} & - & 0.8 (0.78) & 0.59 \\
\textbf{Rater 2} &  & - & 0.65 (0.66)\\
\hline
\end{tabular}
\caption{Pairwise agreement (Cohen's $\kappa$) between raters over all behaviors. A score of 1.0 indicates perfect agreement. Value in parentheses indicate Rater 2 agreement when `uncertain' validation ratings are included.}
\label{tab:post-val}
\end{table}

\section{Conclusion}
\label{sec:concl}

Manually labeling the 24 chat logs used in this work took over 600 hours. Given that the full Perverted-Justice corpus contains 850 chat logs, it would be infeasible to label the entire corpus without the help of automated methods. We find that a transformer-based deep learning approach yields promising results when applied to the detection of online predatory behavior. However, even with training, the agreement between the model and a human annotator is not comparable to the agreement between two human annotators. The model's success in predicting behaviors varies significantly between the categories. Performance is better for more common behaviors, however, we show that the variation is not only due to the number of positive examples in the training data. Extending the prediction task to include multiple messages, increasing the contextual window, boosted detection for certain behaviors. The F1 score for \textit{rapport building}, for example, increased from $61\%$ for single messages to $81\%$ when classification was based on five-messages. Performing post-validation on the automatic classifications allowed us to gain qualitative insight about the model's performance, which may be used to design better prompts and improve performance further. Overall, our results are an encouraging step towards building an automated, psychology-informed model to detect online sexual exploitation.

% \section*{Acknowledgments}
% Darren Cook is funded by \textcolor{red}{Add funding details}. Miri Zilka acknowledges \textcolor{red}{Add funding details}.

\bibliography{CSE_Arxiv}

\begin{thebibliography}{29}
\providecommand{\natexlab}[1]{#1}
\providecommand{\url}[1]{\texttt{#1}}
\expandafter\ifx\csname urlstyle\endcsname\relax
  \providecommand{\doi}[1]{doi: #1}\else
  \providecommand{\doi}{doi: \begingroup \urlstyle{rm}\Url}\fi

\bibitem[Greene-Colozzi et~al.(2020)Greene-Colozzi, Winters, Blasko, and
  Jeglic]{greene2020experiences}
Emily~A Greene-Colozzi, Georgia~M Winters, Brandy Blasko, and Elizabeth~L
  Jeglic.
\newblock Experiences and perceptions of online sexual solicitation and
  grooming of minors: a retrospective report.
\newblock \emph{Journal of child sexual abuse}, 29\penalty0 (7):\penalty0
  836--854, 2020.
\newblock \doi{10.1080/10538712.2020.1801938}.

\bibitem[NSPCC(2021)]{nspcc}
NSPCC.
\newblock Record high number of recorded grooming crimes lead to calls for
  stronger online safety legislation, 2021.
\newblock URL
  \url{https://www.nspcc.org.uk/about-us/news-opinion/2021/online-grooming-record-high}.

\bibitem[Shelton et~al.(2016)Shelton, Eakin, Hoffer, Muirhead, and
  Owens]{shelton2016online}
Joy Shelton, Jennifer Eakin, Tia Hoffer, Yvonne Muirhead, and Jessica Owens.
\newblock Online child sexual exploitation: An investigative analysis of
  offender characteristics and offending behavior.
\newblock \emph{Aggression and violent behavior}, 30:\penalty0 15--23, 2016.
\newblock \doi{10.1016/j.avb.2016.07.002}.

\bibitem[Briggs et~al.(2011)Briggs, Simon, and Simonsen]{briggs2011exploratory}
Peter Briggs, Walter~T Simon, and Stacy Simonsen.
\newblock An exploratory study of internet-initiated sexual offenses and the
  chat room sex offender: Has the internet enabled a new typology of sex
  offender?
\newblock \emph{Sexual Abuse}, 23\penalty0 (1):\penalty0 72--91, 2011.
\newblock \doi{10.1177/1079063210384275}.

\bibitem[Winters et~al.(2017)Winters, Kaylor, and Jeglic]{winters2017sexual}
Georgia~M Winters, Leah~E Kaylor, and Elizabeth~L Jeglic.
\newblock Sexual offenders contacting children online: an examination of
  transcripts of sexual grooming.
\newblock \emph{Journal of sexual aggression}, 23\penalty0 (1):\penalty0
  62--76, 2017.
\newblock \doi{10.1080/13552600.2016.1271146}.

\bibitem[O’Connell(2003)]{o2003typology}
Rachel O’Connell.
\newblock A typology of child cybersexploitation and online grooming practices,
  2003.
\newblock URL
  \url{http://image.guardian.co.uk/sys-files/Society/documents/2003/07/17/Groomingreport.pdf}.

\bibitem[Williams et~al.(2013)Williams, Elliott, and
  Beech]{williams2013identifying}
Rebecca Williams, Ian~A Elliott, and Anthony~R Beech.
\newblock Identifying sexual grooming themes used by internet sex offenders.
\newblock \emph{Deviant Behavior}, 34\penalty0 (2):\penalty0 135--152, 2013.
\newblock \doi{10.1080/01639625.2012.707550}.

\bibitem[Barber and Bettez(2021)]{barber2021exposing}
Connie~S Barber and Silvia~Cristina Bettez.
\newblock Exposing patterns of adult solicitor behaviour: towards a theory of
  control within the cybersexual abuse of youth.
\newblock \emph{European Journal of Information Systems}, 30\penalty0
  (6):\penalty0 591--622, 2021.
\newblock \doi{10.1080/0960085x.2020.1816146}.

\bibitem[Joleby et~al.(2021)Joleby, Lunde, Landstr{\"o}m, and
  Jonsson]{joleby2021offender}
Malin Joleby, Carolina Lunde, Sara Landstr{\"o}m, and Linda~S Jonsson.
\newblock Offender strategies for engaging children in online sexual activity.
\newblock \emph{Child Abuse \& Neglect}, 120:\penalty0 105214, 2021.
\newblock \doi{10.1016/j.chiabu.2021.105214}.

\bibitem[Buckingham and Alali(2020)]{buckingham2020extreme}
Louisa Buckingham and Nusiebah Alali.
\newblock Extreme parallels: a corpus-driven analysis of isis and far-right
  discourse.
\newblock \emph{K{\=o}tuitui: New Zealand Journal of Social Sciences Online},
  15\penalty0 (2):\penalty0 310--331, 2020.
\newblock \doi{10.1080/1177083x.2019.1698623}.

\bibitem[Tausczik and Pennebaker(2010)]{tausczik2010psychological}
Yla~R Tausczik and James~W Pennebaker.
\newblock The psychological meaning of words: Liwc and computerized text
  analysis methods.
\newblock \emph{Journal of language and social psychology}, 29\penalty0
  (1):\penalty0 24--54, 2010.
\newblock \doi{10.1177/0261927x09351676}.

\bibitem[Bogdanova et~al.(2014)Bogdanova, Rosso, and
  Solorio]{bogdanova2014exploring}
Dasha Bogdanova, Paolo Rosso, and Thamar Solorio.
\newblock Exploring high-level features for detecting cyberpedophilia.
\newblock \emph{Computer speech \& language}, 28\penalty0 (1):\penalty0
  108--120, 2014.
\newblock \doi{10.1016/j.csl.2013.04.007}.

\bibitem[Kaur et~al.(2021)Kaur, Singh, and Kaushal]{kaur2021abusive}
Simrat Kaur, Sarbjeet Singh, and Sakshi Kaushal.
\newblock Abusive content detection in online user-generated data: A survey.
\newblock \emph{Procedia Computer Science}, 189:\penalty0 274--281, 2021.
\newblock \doi{10.1016/j.procs.2021.05.098}.

\bibitem[Broome et~al.(2020)Broome, Izura, and Davies]{broome2020psycho}
Laura~Jayne Broome, Cristina Izura, and Jason Davies.
\newblock A psycho-linguistic profile of online grooming conversations: A
  comparative study of prison and police staff considerations.
\newblock \emph{Child Abuse \& Neglect}, 109:\penalty0 104647, 2020.
\newblock \doi{10.1016/j.chiabu.2020.104647}.

\bibitem[Inches and Crestani(2012)]{inches2012overview}
Giacomo Inches and Fabio Crestani.
\newblock Overview of the international sexual predator identification
  competition at pan-2012.
\newblock In \emph{CLEF (Online working notes/labs/workshop)}, volume~30, 2012.

\bibitem[Pendar(2007)]{pendar2007toward}
Nick Pendar.
\newblock Toward spotting the pedophile telling victim from predator in text
  chats.
\newblock In \emph{International Conference on Semantic Computing (ICSC 2007)},
  pages 235--241. IEEE, 2007.
\newblock \doi{10.1109/icsc.2007.32}.

\bibitem[Miah et~al.(2011)Miah, Yearwood, and Kulkarni]{miah2011detection}
Md~Waliur~Rahman Miah, John Yearwood, and Sid Kulkarni.
\newblock Detection of child exploiting chats from a mixed chat dataset as a
  text classification task.
\newblock In \emph{Proceedings of the Australasian Language Technology
  Association Workshop 2011}, pages 157--165, 2011.

\bibitem[Ebrahimi et~al.(2016)Ebrahimi, Suen, and Ormandjieva]{EBRAHIMI201633}
Mohammadreza Ebrahimi, Ching~Y. Suen, and Olga Ormandjieva.
\newblock Detecting predatory conversations in social media by deep
  convolutional neural networks.
\newblock \emph{Digital Investigation}, 18:\penalty0 33--49, 2016.
\newblock ISSN 1742-2876.
\newblock \doi{https://doi.org/10.1016/j.diin.2016.07.001}.
\newblock URL
  \url{https://www.sciencedirect.com/science/article/pii/S1742287616300731}.

\bibitem[Gupta et~al.(2012)Gupta, Kumaraguru, and
  Sureka]{gupta2012characterizing}
Aditi Gupta, Ponnurangam Kumaraguru, and Ashish Sureka.
\newblock Characterizing pedophile conversations on the internet using online
  grooming.
\newblock \emph{arXiv preprint arXiv:1208.4324}, 2012.
\newblock \doi{https://doi.org/10.48550/arXiv.1208.4324}.

\bibitem[Razi et~al.(2021)Razi, Kim, Alsoubai, Stringhini, Solorio,
  De~Choudhury, and Wisniewski]{razi2021human}
Afsaneh Razi, Seunghyun Kim, Ashwaq Alsoubai, Gianluca Stringhini, Thamar
  Solorio, Munmun De~Choudhury, and Pamela~J Wisniewski.
\newblock A human-centered systematic literature review of the computational
  approaches for online sexual risk detection.
\newblock \emph{Proceedings of the ACM on Human-Computer Interaction},
  5\penalty0 (CSCW2):\penalty0 1--38, 2021.
\newblock \doi{10.1145/3479609}.

\bibitem[Perverted-Justice(n.d)]{pjwebsite}
Perverted-Justice.
\newblock \url{http://perverted-justice.com/}, n.d.

\bibitem[Elliott(2017)]{elliott2017self}
Ian~A Elliott.
\newblock A self-regulation model of sexual grooming.
\newblock \emph{Trauma, Violence, \& Abuse}, 18\penalty0 (1):\penalty0 83--97,
  2017.
\newblock \doi{10.1177/1524838015591573}.

\bibitem[Pedregosa et~al.(2011)Pedregosa, Varoquaux, Gramfort, Michel, Thirion,
  Grisel, Blondel, Prettenhofer, Weiss, Dubourg, et~al.]{pedregosa2011scikit}
Fabian Pedregosa, Ga{\"e}l Varoquaux, Alexandre Gramfort, Vincent Michel,
  Bertrand Thirion, Olivier Grisel, Mathieu Blondel, Peter Prettenhofer, Ron
  Weiss, Vincent Dubourg, et~al.
\newblock Scikit-learn: Machine learning in python.
\newblock \emph{Journal of machine learning research}, 12\penalty0
  (Oct):\penalty0 2825--2830, 2011.

\bibitem[Bowman et~al.(2015)Bowman, Angeli, Potts, and
  Manning]{bowman2015large}
Samuel~R Bowman, Gabor Angeli, Christopher Potts, and Christopher~D Manning.
\newblock A large annotated corpus for learning natural language inference.
\newblock \emph{arXiv preprint arXiv:1508.05326}, 2015.
\newblock \doi{10.18653/v1/d15-1075}.

\bibitem[Liu et~al.(2019)Liu, Ott, Goyal, Du, Joshi, Chen, Levy, Lewis,
  Zettlemoyer, and Stoyanov]{liu2019roberta}
Yinhan Liu, Myle Ott, Naman Goyal, Jingfei Du, Mandar Joshi, Danqi Chen, Omer
  Levy, Mike Lewis, Luke Zettlemoyer, and Veselin Stoyanov.
\newblock Roberta: A robustly optimized bert pretraining approach.
\newblock \emph{arXiv preprint arXiv:1907.11692}, 2019.
\newblock \doi{10.48550/arXiv.1907.11692}.

\bibitem[Williams et~al.(2018)Williams, Nangia, and Bowman]{N18-1101}
Adina Williams, Nikita Nangia, and Samuel Bowman.
\newblock A broad-coverage challenge corpus for sentence understanding through
  inference.
\newblock In \emph{Proceedings of the 2018 Conference of the North American
  Chapter of the Association for Computational Linguistics: Human Language
  Technologies, Volume 1 (Long Papers)}, pages 1112--1122. Association for
  Computational Linguistics, 2018.
\newblock \doi{10.18653/v1/n18-1101}.
\newblock URL \url{http://aclweb.org/anthology/N18-1101}.

\bibitem[Wang et~al.(2021)Wang, Fang, Khabsa, Mao, and Ma]{wang2021entailment}
Sinong Wang, Han Fang, Madian Khabsa, Hanzi Mao, and Hao Ma.
\newblock Entailment as few-shot learner.
\newblock \emph{arXiv preprint arXiv:2104.14690}, 2021.
\newblock \doi{10.48550/arXiv.2104.14690}.

\bibitem[Cohen(1960)]{cohen1960coefficient}
Jacob Cohen.
\newblock A coefficient of agreement for nominal scales.
\newblock \emph{Educational and psychological measurement}, 20\penalty0
  (1):\penalty0 37--46, 1960.
\newblock \doi{10.1177/001316446002000104}.

\bibitem[McHugh(2012)]{mchugh2012interrater}
Mary~L McHugh.
\newblock Interrater reliability: the kappa statistic.
\newblock \emph{Biochemia medica}, 22\penalty0 (3):\penalty0 276--282, 2012.
\newblock \doi{10.11613/bm.2012.031}.

\end{thebibliography}
\bibliographystyle{unsrtnat}

\clearpage 

\appendix
% \beginsupplement

% Content label descriptions.
\section{Content Label Descriptions}\label{app-label-desc}
\subsection*{Communication/Coordination}
\paragraph{Description: } ``Communication Coordination'' is used to start and maintain communication as offenders: (i) exchange and clarify information with their intended victim, (ii) present reason/excuses, (iii) assess the level of engagement of the victim, (iv) find new ways to communicate (i.e., media exchange), (v) strategically use humour or linguistic fillers (i.e., ``lol'', ``hehe''), and (vi) redirect the flow of conversation. One of the offenders' main purposes of this category is to maximize gain and potentially minimize time spent on non-compliant victims.

\subsection*{Rapport Building}
\paragraph{Description: } Offenders use positive behavior to mimic romantic relationships, making it easier to introduce sexual topics \citep{elliott2017self}. Offenders use ‘Rapport’ to infiltrate victims’ offline/online social and emotional life to create an illusion of exclusivity, reinforcing the offender as a trusted other. This is achieved through compliments/sweet talk, showing interest, and shared experiences. This special connection or bond is usually created in a short amount of time through excessive saturation and exposure to constant positive statements \citep{elliott2017self}.

\subsection*{Control/Regulation}
\paragraph{Description: } Control/Regulation’ occurs when offenders use power to direct the flow of communication by influencing or directing the victim's behavior. Controlling the conversation can occur through subtle (e.g., illusion of control, rhetorical questions, checking for willingness to engage, or permissive behavior) or direct strategies (e.g., making demands, persistence, use of coercion). 
Offenders may attempt to take control of the conversation through patronizing language, persistence, frequently checking for engagement, making demands, or by asking questions that give the illusion of consent - giving the impression that victims have control over what happens during an exchange.

\subsection*{Challenges}
\paragraph{Description: } An offender may challenge a victim when opposing motivations appear. As a result, confrontation ensues directly (e.g., offence, control, aggression) or indirectly (e.g., joke, mockery, irony). Offenders often challenge the victim as a way of authenticating identity, or to exert more control.

\subsection*{Negotiation}
\paragraph{Description: } ``Negotiation'' can occur at any time during the exchange and is the process where offenders attempt to make decisions, compromise, incentivize continued interaction, or reach goal achievement (e.g., confirming a plan to meet). ``Negotiations'' can be brief or extensive depending on what goals the offender is trying to achieve. Incentives are particularly important when negotiating goals, and can be either financial or emotional.

\subsection*{Use of Emotion}
\paragraph{Description: } Offenders use emotive language to  manipulate the victim's emotions in order to influence their behavior. ``Use of emotions'' can be positive or negative, and include sub-behaviors such as manipulation, expressing empathy, guilt tripping, vilifying third parties, offering reassurance, or by playing the victim. Offenders may employ positive strategies to isolate victims, and use negative emotions increase compliance.

\subsection*{Testing Boundaries}
\paragraph{Description: } ``Testing Boundaries'’ determines whether the conversation continues or ends. Offenders seek to test boundaries directly or indirectly to determine whether it is possible to desensitize victims through exposure to sexual topics \citep{elliott2017self}. 

\subsection*{Use of Sexual Topics}
\paragraph{Description: } Offenders intentionally use sex to desensitize victims. This is done by directing conversation toward the victim's prior sexual experiences, discussing fantasies, use of explicit language, determining sexual preferences, suggesting media production, alluding to traveling for sex, and acting as a sexual mentor.

\subsection*{Mitigation}
\paragraph{Description: } ``Mitigation'' is a strategy that aims to soften or downgrade the intensity or seriousness of what is being expressed to convince the victim to participate. Offenders may use this technique in an attempt to normalize the sexual exchange by lessening the idea of harm or criminality. Specific sub-behaviors include indirectly stating a sexual preference for children, implicating oneself in a previous crime, normalizing sexual conversations, or discussing differences in age. Normalization occurs by talking about sex often without reservation and is the process of desensitizing the victim to sexual topics or acts.

\subsection*{Encouraging}
\paragraph{Description: } Offenders use encouragement to comply with the victims requests, or to show support by acting as a mentor/trusted other.

\subsection*{Risk Management}
\paragraph{Description: } ``Risk Management'' occurs when offenders assess risk and take steps to prevent discovery. This may be through incentivizing secrecy using emotional manipulation, asking the decoy to delete messages/images, enquiring after third parties (e.g., the location of parents), acknowledge previous wrongdoing, and discussing the consequences of getting caught.

\clearpage

\onecolumn

\section{Hyperparameter Tuning}\label{hyp-app}
\begin{table*}[h!]
\centering
\begin{tabular}{llc}
\hline
\textbf{Algorithm} & \textbf{Parameter} & \textbf{Value Range}\\
\hline
\multirow{2}{*}{Random Forest}&n\_estimators&$50$, $100$*, $150$\\
& max\_depth&$10$, $50$, $100$\vspace{0.3cm}\\
\multirow{2}{*}{Logistic Regression}&solver&liblinear, saga\\
& penalty&\textit{l1}, \textit{l2}*\vspace{0.3cm}\\
\multirow{2}{*}{Support Vector Machine}&C&$0.1$, $0.5$, $1$*\\
& kernel&linear, poly, rbf*\vspace{0.3cm}\\
\multirow{2}{*}{Naive Bayes}&alpha&$0.0$, $1.0$*\\
& fit\_prior&True*, False\vspace{0.3cm}\\
\hline
\end{tabular}
\caption{Range of settings evaluated for each supervised algorithm. As GridSearch is an exhaustive optimization technique, effort was taken to minimize the number of models tested. Algorithms were implemented via \texttt{Scikit-learn}. * indicates the default setting.}
\label{tab:hyp-vals}
\end{table*}

\clearpage

\section{Precision, Recall, and Accuracy Plots for Full-Shot Single-Classification Models}

\begin{figure*}[h]
\centerline{\includegraphics[scale=0.55]{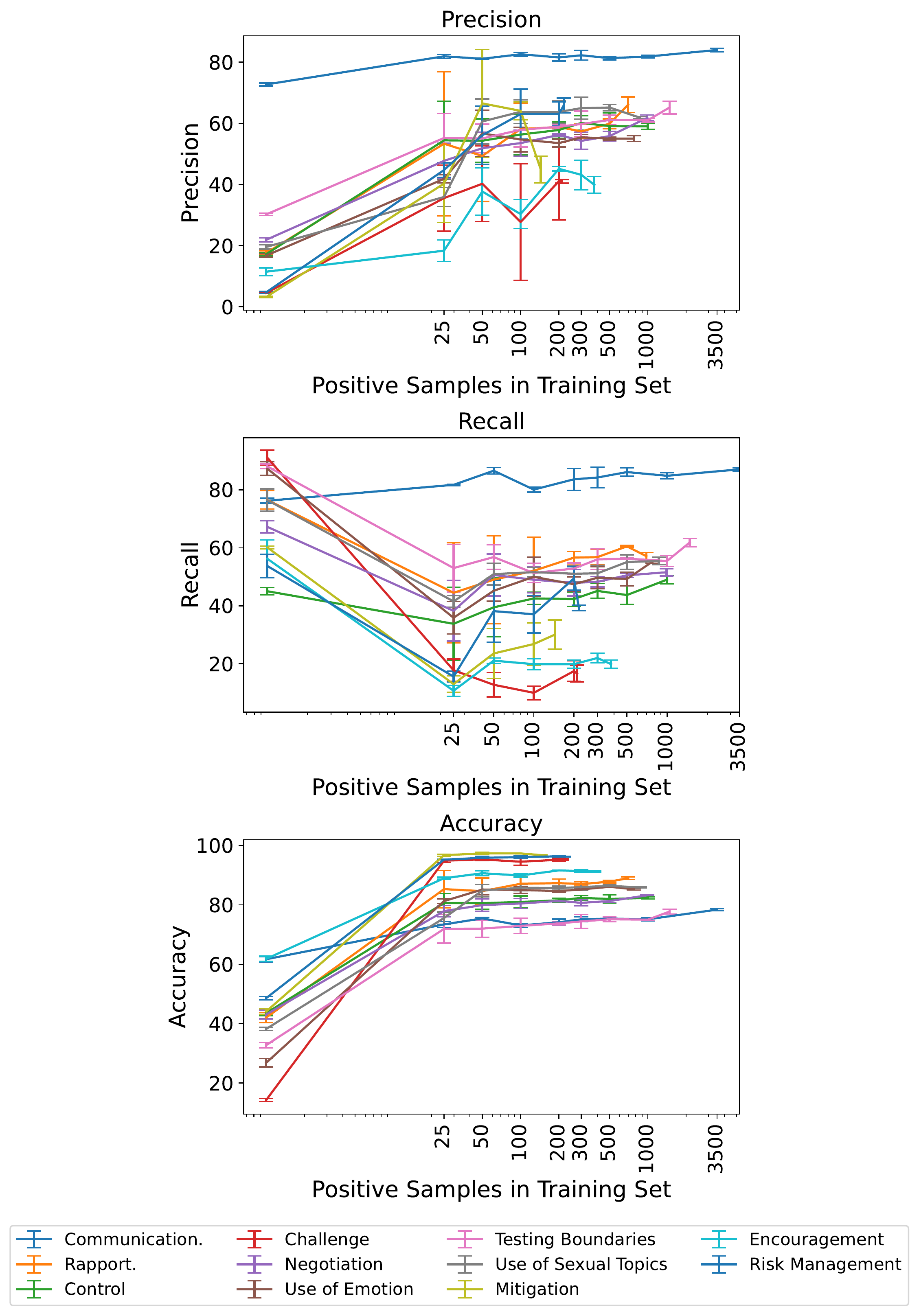}}
\caption{Precision, recall, and accuracy of single message classification as the number of positive training examples increases (up to full training set). Error bars are standard  deviations.}
\label{fewshot-perf-change-si}
\end{figure*}

\clearpage

\section{Inter-Rater Agreement Scores for Each Behavior}\label{app:agree}
\begin{table*}[h!]
\centering
\begin{tabular}{lcccc}
\hline
\textbf{Behavior} & \textbf{Rater 1/Rater 2} & \textbf{Rater 1/Model} & \textbf{Rater 2/Model} & \textbf{Overall}\\
\hline
Communication/Coordination&$0.66$ ($0.65$)&$0.53$&$\mathbf{0.71}$ ($0.72$)&$0.63$ ($0.62$)\\
Rapport Building&$\mathbf{0.73}$ ($0.7$)&$0.63$&$0.62$ ($0.65$)&$0.66$ ($0.65$)\\
Control&$\mathbf{0.62}$ ($0.6$)&$0.52$&$0.55$ ($0.59$)&$0.56$ ($0.57$)\\
Challenge&$\mathbf{0.89}$ ($0.89$)&$0.46$&$0.51$ ($0.51$)&$0.62$ ($0.62$)\\
Negotiation&$\mathbf{0.95}$ ($0.93$)&$0.67$&$0.65$ ($0.67$)&$0.76$ ($0.76$)\\
Use of Emotions&$\mathbf{0.78}$ ($0.74$)&$0.6$&$0.77$ ($0.79$)&$0.72$ ($0.7$)\\
Testing Boundaries&$\mathbf{0.88}$ ($0.84$)&$0.5$&$0.52$ ($0.53$)&$0.63$ ($0.62$)\\
Use of Sexual Topics&$\mathbf{0.74}$ ($0.73$)&$0.6$&$0.64$ ($0.65$)&$0.66$ ($0.66$)\\
Mitigation&$\mathbf{0.91}$ ($0.91$)&$0.52$&$0.52$ ($0.52$)&$0.65$ ($0.65$)\\
Encouragement&$0.66$ ($0.66$)&$0.61$&$\mathbf{0.84}$ ($0.84$)&$0.7$ ($0.71$)\\
Risk Management&$\mathbf{0.88}$ ($0.85$)&$0.67$&$0.67$ ($0.68$)&$0.74$ ($0.72$)\\
\hline
Total &$\mathbf{0.8}$ ($0.78$)&$0.59$&$0.65$ ($0.66$)&$0.68$ ($0.67$)\\
\hline
\end{tabular}
\caption{Pairwise Cohen's $\kappa$ per behavior label. Overall score is the mean of the pairwise scores. $1.0$ indicates perfect agreement; 0.0 indicates no agreement. Values greater than $\kappa=0.6$ are generally indicative of ``moderate to strong'' agreement, as per guidelines in \citep{mchugh2012interrater}.  $\kappa$ in \textbf{bold} indicates largest pairwise agreement per behavior label. \textit{Rater 1} is the actual labels used in training (i.e., ground-truth), \textit{Model} refers to the predictions generated by the best performing single-classification model for each behavior, and \textit{Rater 2} refers to manual ratings generated during post-validation. Values in parentheses indicate $\kappa$ when ``unsure'' ratings given by \textit{Rater 2} are included as positive agreement. }
\label{tab:kappa-behavior}
\end{table*}

\clearpage

\section{Full Model Results}\label{full-results}

\begin{table*}[h]
\centering
\begin{tabular}{lcccc}
\hline
\textbf{Model} & \textbf{Accuracy} & \textbf{Precision} & \textbf{Recall} & \textbf{F1}\\
\hline
\vspace{0.1cm}\\
\textit{Supervised - Bag of Words}\\
Random Forest & $73.8$ ($\pm0.26$) & $74.28$ ($\pm0.11$) & $98.18$ ($\pm0.8$) & $84.57$ ($\pm0.23$) \\
Logistic Regression & $73.73$ ($\pm0.2$) & $75.5$ ($\pm0.11$) & $94.85$ ($\pm0.61$) & $84.08$ ($\pm0.18$) \\
Support Vector Machine & $74.59$ ($\pm0.62$) & $75.41$ ($\pm0.26$) & $96.84$ ($\pm0.61$) & $84.79$ ($\pm0.39$) \\
Naive Bayes & $74.47$ ($\pm1.35$) & $76.5$ ($\pm0.58$) & $93.95$ ($\pm1.31$) & $84.33$ ($\pm0.88$) \\

\vspace{0.3cm}\\
\textit{Transformer}\\
0-shot & $61.7$ ($\pm0.87$) & $72.71$ ($\pm0.46$)& $76.25$ ($\pm0.84$)& $74.44$ ($\pm0.64$)\\
25-shot & $73.43$ ($\pm1.01$) & $81.91$ ($\pm0.68$)& $81.74$ ($\pm0.24$)& $81.81$ ($\pm0.98$)\\
50-shot & $75.45$ ($\pm0.36$) & $81.11$ ($\pm0.28$)& $86.61$ ($\pm1.13$)& $83.77$ ($\pm0.38$)\\
100-shot & $73.09$ ($\pm0.69$)& $82.59$ ($\pm0.61$)& $80.09$ ($\pm0.85$)& $81.32$($\pm0.51$)\\
200-shot & $74.15$ ($\pm.1.08$) & $81.54$ ($\pm1.2$)& $83.65$ ($\pm3.8$)& $82.53$ ($\pm1.28$)\\
300-shot & $75.13$ ($\pm0.84$) & $82.25$ ($\pm1.58$)& $84.26$ ($\pm3.54$)& $83.19$ ($\pm1.02$)\\
500-shot & $75.38$ ($\pm0.37$) & $81.31$ ($\pm0.58$)& $86.14$ ($\pm1.41$)& $83.65$ ($\pm0.41$)\\
1000-shot & $75.15$ ($\pm0.57$) & $81.84$ ($\pm0.44$)& $84.86$ ($\pm1.06$)& $83.32$ ($\pm0.47$)\\
\textbf{Full-shot (3466)} & $78.4$ ($\pm0.35$) & $83.98$ ($\pm0.55$) & $87.08$ ($\pm0.53$) & $85.5$ ($\pm0.21$)  \\
\vspace{0.1cm}\\
\hline
\end{tabular}
\caption{Performance of classification models for \textbf{Communication/Coordination}. Full-shot value refers to the number of positive examples present in the full training set. Best performing model is in bold.}
\label{tab:app-communication-res}
\end{table*}

\begin{table*}[ht]
\centering
\begin{tabular}{lcccc}
\hline
\textbf{Model} & \textbf{Accuracy} & \textbf{Precision} & \textbf{Recall} & \textbf{F1}\\
\hline
\vspace{0.1cm}\\
\textit{Supervised - Bag of Words}\\
Random Forest & $87.09$ ($\pm0.52$) & $71.26$ ($\pm6.61$) & $25.57$ ($\pm2.24$) & $37.55$ ($\pm2.51$) \\
Logistic Regression & $87.16$ ($\pm0.07$) & $72.28$ ($\pm3.17$) & $25.41$ ($\pm1.96$) & $37.52$ ($\pm1.67$) \\
Support Vector Machine & $86.32$ ($\pm0.31$) & $69.8$ ($\pm1.37$) & $17.8$ ($\pm4.07$) & $28.18$ ($\pm5.25$) \\
Naive Bayes & $86.54$ ($\pm0.19$) & $72.17$ ($\pm3.38$) & $18.77$ ($\pm0.28$) & $29.78$ ($\pm0.4$) \\

\vspace{0.3cm}\\
\textit{Transformer}\\
% 0-shot & $42.51$ & $17.26$ & $73.3$ & $27.94$\\
% 25-shot & $81.03$ & $36.36$ & $33.01$ & $34.61$ \\
% 50-shot & $81.92$ & $40.3$ & $39.32$ & $39.8$\\
% 100-shot & $84.35$ & $48.3$ & $41.26$ & $44.5$\\
% 200-shot & $86.2$ & $54.92$ & $51.46$ & $53.13$\\
% 300-shot & $87.6$ & $59.79$ & $56.31$ & $58.0$\\
% 500-shot & $87.6$ & $59.41$ & $58.25$ & $58.82$\\
% 1000-shot & $88.56$ & $64.91$ & $53.88$ & $58.89$ \\
% Full-shot & $89.67$& $67.93$ & $60.68$ & $64.1$ \\
0-shot & $42.07$ ($\pm1.67$) & $17.63$ ($\pm0.73$)& $76.54$ ($\pm3.16$)& $28.66$ ($\pm1.15$)\\
25-shot & $85.31$ ($\pm6.27$) & $53.34$ ($\pm23.53$)  & $44.5$ ($\pm17.21$)  & $48.45$ ($\pm19.82$) \\
50-shot & $84.63$ ($\pm4.45$) & $49.33$ ($\pm14.86$) & $49.03$ ($\pm15.13$) & $49.18$($\pm15.0$) \\
100-shot & $87.11$ ($\pm2.32$) & $58.15$ ($\pm8.51$) & $52.1$ ($\pm11.53$) & $54.86$($\pm10.08$) \\
200-shot & $87.33$ ($\pm1.36$) & $58.79$ ($\pm4.96$) & $56.63$ ($\pm2.19$) & $57.67$($\pm3.53$) \\
300-shot & $87.01$ ($\pm0.71$) & $57.36$ ($\pm2.36$) & $56.8$ ($\pm2.7$) & $57.07$($\pm2.41$) \\
500-shot & $87.82$ ($\pm0.41$) & $59.87$ ($\pm1.59$) & $60.52$ ($\pm0.28$) & $60.18$($\pm0.8$) \\
% 1000-shot & $88.24$ ($\pm0.09$) & $62.03$ ($\pm0.78$) & $58.58$ ($\pm3.77$) & $60.19$ ($\pm0.17$) \\
1000-shot* \\
Full-shot (721) & $89.0$ ($\pm0.45$) & $66.08$ ($\pm2.58$) & $57.12$ ($\pm1.22$) & $61.24$ ($\pm0.59$) \\

\vspace{0.1cm}\\
\textit{Transformer - Multi-Message Input}\\
Full-shot 3 Msg & $91.48$ ($\pm6.31$) & $80.47$ ($\pm15.23$) & $77.97$ ($\pm17.94$) & $79.16$ ($\pm16.58$) \\
\textbf{Full-shot 5 Msg} & $89.79$ ($\pm6.34$) & $82.95$ ($\pm9.27$) & $80.05$ ($\pm15.71$) & $81.33$ ($\pm12.58$) \\
\hline
\end{tabular}
\caption{Performance of classification models for \textbf{Rapport Building}. Full-shot value refers to the number of positive examples present in the full training set. Best performing model is in bold.}
{\raggedright * The number exceeds the total number of positive examples in the dataset and therefore results are analogous to full-shot \par}
\label{tab:app-rapport-res}
\end{table*}

\begin{table*}[ht]
\centering
\begin{tabular}{lcccc}
\hline
\textbf{Model} & \textbf{Accuracy} & \textbf{Precision} & \textbf{Recall} & \textbf{F1}\\
\hline
\vspace{0.1cm}\\
\textit{Supervised - Bag of Words}\\
Random Forest & $79.75$ ($\pm0.54$) & $53.84$ ($\pm3.61$) & $19.03$ ($\pm1.34$) & $28.12$ ($\pm1.95$)\\
Logistic Regression & $79.85$ ($\pm1.0$) & $56.08$ ($\pm8.61$) & $15.25$ ($\pm2.16$) & $23.96$ ($\pm3.38$) \\
Support Vector Machine & $80.17$ ($\pm0.74$) & $59.35$ ($\pm6.21$) & $14.89$ ($\pm3.15$) & $23.74$ ($\pm4.37$) \\
Naive Bayes& $79.41$ ($\pm0.48$) & $52.87$ ($\pm5.14$) & $11.35$ ($\pm1.42$) & $18.64$ ($\pm2.08$)

\vspace{0.3cm}\\
\textit{Transformer}\\
0-shot & $43.57$ ($\pm0.86$) & $17.24$ ($\pm0.51$) & $45.04$ ($\pm1.23$) & $24.94$ ($\pm0.71$)\\
25-shot & $80.64$ ($\pm3.13$) & $54.44$ ($\pm12.75$) & $33.81$ ($\pm12.56$) & $41.53$ ($\pm13.31$)\\
50-shot & $80.59$ ($\pm2.11$) & $54.32$ ($\pm7.12$) & $39.48$ ($\pm10.13$) & $45.5$ ($\pm8.72$)\\
100-shot & $80.98$ ($\pm2.01$) & $56.33$ ($\pm6.7$) & $42.55$ ($\pm2.16$) & $48.3$ ($\pm2.32$)\\
200-shot & $81.55$ ($\pm0.7$) & $57.82$ ($\pm2.7$) & $42.35$ ($\pm2.49$) & $48.89$ ($\pm1.75$)\\
300-shot & $82.36$ ($\pm0.78$) & $60.13$ ($\pm2.38$) & $45.15$ ($\pm2.56$) & $51.57$ ($\pm2.52$)\\
500-shot & $81.99$ ($\pm1.34$) & $59.11$ ($\pm4.38$) & $43.73$ ($\pm3.22$) & $50.27$ ($\pm3.68$) \\
% 1000-shot &  $82.46$ ($\pm0.41$) & $60.79$ ($\pm0.58$) & $44.21$ ($\pm3.55$) & $51.15$ ($\pm2.58$)\\
1000-shot* \\
Full-shot (986) & $82.36$ ($\pm0.37$) & $59.2$ ($\pm1.0$) & $49.05$ ($\pm1.43$) & $53.65$ ($\pm1.2$)\\
\vspace{0.1cm}\\
\textit{Transformer - Multi-Message Input}\\
Full-shot 3 Msg &  $76.53$ ($\pm0.33$) & $59.12$ ($\pm1.1$) & $41.52$ ($\pm0.67$) & $48.78$ ($\pm0.48$)\\
\textbf{Full-shot 5 Msg} &  $71.32$ ($\pm0.85$) & $61.47$ ($\pm1.74$) & $59.47$ ($\pm1.39$) & $60.43$ ($\pm0.15$)\\
\hline
\end{tabular}
\caption{Performance of classification models for \textbf{Control}. Full-shot value refers to the number of positive examples present in the full training set. Best performing model is in bold.}
{\raggedright * The number exceeds the total number of positive examples in the dataset and therefore results are analogous to full-shot \par}
\label{tab:app-control-res}
\end{table*}

\begin{table*}[ht]
\centering
\begin{tabular}{lcccc}
\hline
\textbf{Model} & \textbf{Accuracy} & \textbf{Precision} & \textbf{Recall} & \textbf{F1}\\
\hline
\vspace{0.1cm}\\
\textit{Supervised - Bag of Words}\\
Random Forest & $95.43$ ($\pm0.01$) & $20.0$ ($\pm20.0$) & $1.67$ ($\pm1.67$) & $3.08$ ($\pm3.08$)\\
Logistic Regression & $95.42$ ($\pm0.01$) & $13.33$ ($\pm23.1$) & $1.11$ ($\pm1.93$) & $2.05$ ($\pm3.55$) \\
Support Vector Machine & $95.55$ ($\pm0.01$) & $0.0$ ($\pm0.0$) & $0.0$ ($\pm0.0$) & $0.0$ ($\pm0.0$) \\
Naive Bayes& $95.55$ ($\pm0.01$) & $50.5$ ($\pm50.5$) & $1.11$ ($\pm0.96$) & $2.17$ ($\pm1.88$)

\vspace{0.3cm}\\
\textit{Transformer}\\
0-shot & $14.29$ ($\pm0.54$) & $4.52$ ($\pm0.09$) & $91.11$ ($\pm2.55$) & $8.6$ ($\pm0.18$)\\
25-shot & $94.88$ ($\pm0.57$) & $35.56$ ($\pm10.81$) & $17.78$ ($\pm3.85$) & $23.66$ ($\pm5.78$)\\
50-shot & $95.3$ ($\pm0.35$) & $40.25$ ($\pm12.39$) & $12.78$ ($\pm4.19$) & $19.39$ ($\pm6.25$)\\
100-shot & $94.54$ ($\pm1.15$) & $27.73$ ($\pm19.01$) & $10.0$ ($\pm2.36$) & $14.35$ ($\pm5.41$)\\
200-shot & $95.2$ ($\pm0.52$) & $41.07$ ($\pm12.63$) & $17.5$ ($\pm3.54$) & $24.51$ ($\pm5.74$)\\
300-shot* & \\
500-shot* &  \\
1000-shot* &  \\
Full-shot (211) & $95.25$ ($\pm0.23$) & $41.04$ ($\pm0.62$) & $16.67$ ($\pm2.89$) & $23.7$ ($\pm3.93$)\\
\vspace{0.1cm}\\
\textit{Transformer - Multi-Message Input}\\
Full-shot 3 Msg & $91.96$ ($\pm0.24$) & $15.53$ ($\pm1.79$) & $5.36$ ($\pm1.79$) & $17.97$ ($\pm2.67$)\\
\textbf{Full-shot 5 Msg} & $88.31$ ($\pm1.12$) & $36.14$ ($\pm1.68$) & $24.18$ ($\pm4.08$) & $28.9$ ($\pm3.51$)\\
\hline
\end{tabular}
\caption{Performance of classification models for \textbf{Challenge}. Full-shot value refers to the number of positive examples present in the full training set. Best performing model is in bold.}
{\raggedright * The number exceeds the total number of positive examples in the dataset and therefore results are analogous to full-shot \par}
\label{tab:app-challenge-res}
\end{table*}

\begin{table*}[ht]
\centering
\begin{tabular}{lcccc}
\hline
\textbf{Model} & \textbf{Accuracy} & \textbf{Precision} & \textbf{Recall} & \textbf{F1}\\
\hline
\vspace{0.1cm}\\
\textit{Supervised - Bag of Words}\\
Random Forest & $80.57$ ($\pm1.08$) & $59.57$ ($\pm6.74$) & $21.2$ ($\pm4.95$) & $31.14$ ($\pm6.18$)\\
Logistic Regression & $81.45$ ($\pm1.08$) & $67.55$ ($\pm9.62$) & $22.5$ ($\pm1.14$) & $33.65$ ($\pm1.76$) \\
Support Vector Machine & $80.57$ ($\pm0.66$) & $65.05$ ($\pm7.93$) & $13.43$ ($\pm3.08$) & $22.3$ ($\pm4.46$)\\
Naive Bayes & $81.35$ ($\pm0.7$) & $66.58$ ($\pm7.39$) & $22.5$ ($\pm3.19$) & $33.42$ ($\pm3.04$)

\vspace{0.3cm}\\
\textit{Transformer}\\
% 0-shot & $41.85$ & $21.21$ & $65.72$ & $32.07$ \\
% 25-shot & $76.67$ & $39.62$ & $22.26$ & $28.51$ \\
% 50-shot & $76.46$ & $41.0$ & $28.98$ & $33.96$\\
% 100-shot & $77.79$ & $45.91$ & $35.69$ & $40.16$\\
% 200-shot & $80.44$ & $53.15$ & $53.71$ & $53.43$ \\
% 300-shot & $81.48$ & $56.2$ & $51.24$ & $53.6$\\
% 500-shot & $81.11$ & $55.29$ & $49.82$ & $52.42$ \\
% 1000-shot & $81.03$ & $54.89$ & $51.59$ & $53.19$ \\
% Full-shot & $80.96$ & $54.39$ & $54.77$ & $54.58$ \\
0-shot & $43.1$ ($\pm1.47$) & $21.92$ ($\pm0.62$) & $67.26$ ($\pm2.07$) & $33.06$ ($\pm0.86$)\\
25-shot & $77.91$ ($\pm1.9$) & $47.73$ ($\pm5.8$) & $38.28$ ($\pm10.45$) & $41.49$ ($\pm4.78$)\\
50-shot & $79.93$ ($\pm2.11$) & $51.87$ ($\pm5.23$) & $50.65$ ($\pm7.21$) & $51.22$ ($\pm6.14$)\\
100-shot & $80.47$ ($\pm1.65$) & $53.54$ ($\pm4.26$) & $49.0$ ($\pm4.45$) & $51.15$ ($\pm4.22$)\\
200-shot & $81.33$ ($\pm0.27$) & $56.21$ ($\pm0.36$) & $47.94$ ($\pm4.54$) & $51.67$ ($\pm2.76$)\\
300-shot & $80.69$ ($\pm1.01$) & $54.27$ ($\pm2.81$) & $48.29$ ($\pm1.78$) & $51.1$ ($\pm2.16$)\\
500-shot & $81.3$ ($\pm0.52$) & $55.83$ ($\pm1.58$) & $50.53$ ($\pm0.71$) & $53.03$ ($\pm0.55$) \\
% 1000-shot & $81.03$ ($\pm1.08$) & $54.89$ ($\pm0.71$) & $51.59$ ($\pm2.1$) & $53.19$ ($\pm0.82$) \\
1000-shot* \\
Full-shot (991) & $83.22$ ($\pm0.15$) & $61.8$ ($\pm0.92$) & $51.59$ ($\pm1.27$) & $56.22$ ($\pm0.39$) \\
\vspace{0.1cm}\\
\textit{Transformer - Multi-Message Input}\\
Full-shot 3 Msg & $80.43$ ($\pm1.13$) & $63.31$ ($\pm2.41$) & $61.99$ ($\pm1.41$) & $62.64$ ($\pm1.84$) \\
\textbf{Full-shot 5 Msg} & $74.74$ ($\pm1.96$) & $62.85$ ($\pm2.93$) & $66.67$ ($\pm1.71$) & $64.7$ ($\pm2.31$) \\
\hline
\end{tabular}
\caption{Performance of classification models for \textbf{Negotiation}. Full-shot value refers to the number of positive examples present in the full training set. Best performing model is in bold.}
{\raggedright * The number exceeds the total number of positive examples in the dataset and therefore results are analogous to full-shot \par}
\label{tab:app-negotiation-res}
\end{table*}

\begin{table*}[ht]
\centering
\begin{tabular}{lcccc}
\hline
\textbf{Model} & \textbf{Accuracy} & \textbf{Precision} & \textbf{Recall} & \textbf{F1}\\
\hline
\vspace{0.1cm}\\
\textit{Supervised - Bag of Words}\\
Random Forest & $84.82$ ($\pm0.69$) & $66.26$ ($\pm8.67$) & $14.87$ ($\pm2.95$) & $24.25$ ($\pm4.4$)\\
Logistic Regression & $84.38$ ($\pm0.53$) & $57.47$ ($\pm0.52$) & $17.71$ ($\pm2.27$) & $27.07$ ($\pm3.17$) \\
Support Vector Machine & $84.4$ ($\pm0.55$) & $68.5$ ($\pm11.75$) & $8.56$ ($\pm2.51$) & $15.2$ ($\pm4.27$)\\
Naive Bayes & $84.77$ ($\pm0.31$) & $65.59$ ($\pm4.21$) & $14.87$ ($\pm0.9$) & $24.24$ ($\pm1.47$)

\vspace{0.3cm}\\
\textit{Transformer}\\
% 0-shot & $26.79$ & $16.52$ & $85.59$ & $27.7$ \\
% 25-shot & $80.52$ & $37.79$ & $29.28$ & $33.0$ \\
% 50-shot & $84.43$ & $53.5$ & $37.84$ & $44.33$\\
% 100-shot & $85.09$ & $56.33$ & $40.09$ & $46.84$ \\
% 200-shot & $83.76$ & $50.49$ & $46.85$ & $48.6$ \\
% 300-shot & $84.65$ & $53.57$ & $47.3$ & $50.24$\\
% 500-shot &  $85.31$ & $56.5$ & $45.05$ & $50.13$\\
% 1000-shot & $85.24$ & $55.79$ & $47.75$ & $51.46$ \\
% Full-shot & $85.09$ & $55.32$ & $46.85$ & $50.73$ \\
0-shot & $26.83$ ($\pm1.4$) & $16.77$ ($\pm0.55$) & $87.34$ ($\pm2.39$) & $28.13$ ($\pm0.89$)\\
25-shot & $81.28$ ($\pm0.81$) & $41.86$ ($\pm1.31$) & $35.89$ ($\pm5.63$) & $38.42$ ($\pm2.82$)\\
50-shot & $85.17$ ($\pm1.76$) & $56.69$ ($\pm7.54$) & $45.2$ ($\pm3.62$) & $50.02$ ($\pm2.98$)\\
100-shot & $85.02$ ($\pm1.11$) & $54.73$ ($\pm4.03$) & $50.0$ ($\pm6.76$) & $52.12$ ($\pm4.62$)\\
200-shot & $84.65$ ($\pm0.38$) & $53.54$ ($\pm1.24$) & $47.45$ ($\pm2.56$) & $50.3$ ($\pm1.97$)\\
300-shot & $85.22$ ($\pm0.37$) & $55.41$ ($\pm0.94$) & $49.7$ ($\pm3.83$) & $52.37$ ($\pm2.5$)\\
500-shot & $85.09$ ($\pm0.2$) & $55.02$ ($\pm0.47$) & $49.25$ ($\pm2.31$) & $51.96$ ($\pm1.49$) \\
% 1000-shot & $84.82$ ($\pm0.24$) & $54.1$ ($\pm1.06$) & $48.95$ ($\pm2.03$) & $51.37$ ($\pm0.75$) \\
1000-shot* \\
Full-shot (777) & $85.27$ ($\pm0.26$) & $55.02$ ($\pm0.91$) & $55.26$ ($\pm0.52$) & $55.13$ ($\pm0.28$) \\
\vspace{0.1cm}\\
\textit{Transformer - Multi-Message Input}\\
Full-shot 3 Msg & $78.89$ ($\pm1.03$) & $52.16$ ($\pm2.72$) & $44.62$ ($\pm2.61$) & $48.09$ ($\pm2.67$) \\
\textbf{Full-shot 5 Msg} & $73.64$ ($\pm0.85$) & $58.21$ ($\pm1.44$) & $53.13$ ($\pm1.65$) & $55.55$ ($\pm1.56$) \\
\hline
\end{tabular}
\caption{Performance of classification models for \textbf{Use of Emotions}. Full-shot value refers to the number of positive examples present in the full training set. Best performing model is in bold.}
{\raggedright * The number exceeds the total number of positive examples in the dataset and therefore results are analogous to full-shot \par}
\label{tab:app-using-emotions-res}
\end{table*}

\begin{table*}[ht]
\centering
\begin{tabular}{lcccc}
\hline
\textbf{Model} & \textbf{Accuracy} & \textbf{Precision} & \textbf{Recall} & \textbf{F1}\\
\hline
\vspace{0.1cm}\\
\textit{Supervised - Bag of Words}\\
Random Forest & $72.87$ ($\pm1.39$) & $63.03$ ($\pm4.09$) & $31.52$ ($\pm2.82$) & $42.02$ ($\pm3.41$)\\
Logistic Regression & $72.96$ ($\pm1.32$) & $62.47$ ($\pm4.2$) & $33.73$ ($\pm1.66$) & $43.79$ ($\pm2.31$) \\
Support Vector Machine & $73.31$ ($\pm0.75$) & $63.39$ ($\pm2.71$) & $34.44$ ($\pm0.55$) & $44.62$ ($\pm1.0$)\\
Naive Bayes & $73.65$ ($\pm0.7$) & $63.34$ ($\pm2.22$) & $37.12$ ($\pm0.47$) & $46.8$ ($\pm0.95$)

\vspace{0.3cm}\\
\textit{Transformer}\\
% 0-shot & $32.1$ & $29.88$ & $87.23$ & $44.51$ \\
% 25-shot & $74.69$ & $57.91$ & $69.27$ & $63.08$ \\
% 50-shot & $73.51$ & $55.98$ & $76.36$ & $64.28$ \\
% 100-shot & $75.2$ & $58.9$ & $68.09$ & $63.16$ \\
% 200-shot & $74.02$ & $56.79$ & $70.21$ & $62.79$ \\
% 300-shot & $76.09$ & $61.33$ & $63.36$ & $62.33$\\
% 500-shot & $75.72$ & $61.75$ & $58.39$ & $60.02$ \\
% 1000-shot &  $75.5$ & $60.18$ & $63.59$ & $61.84$\\
% Full-shot & $77.12$ & $62.2$ & $68.09$ & $65.01$ \\

0-shot & $32.77$ ($\pm0.85$) & $30.22$ ($\pm0.39$) & $88.1$ ($\pm0.83$) & $45.0$ ($\pm0.54$)\\
25-shot & $71.93$ ($\pm4.84$) & $55.22$ ($\pm8.07$) & $53.03$ ($\pm8.18$) & $54.1$ ($\pm8.1$)\\
50-shot & $72.01$ ($\pm2.87$) & $55.05$ ($\pm4.66$) & $56.82$ ($\pm4.29$) & $55.91$ ($\pm4.39$)\\
100-shot & $72.92$ ($\pm2.61$) & $57.88$ ($\pm5.58$) & $51.3$ ($\pm3.34$) & $54.22$ ($\pm2.34$)\\
200-shot & $73.78$ ($\pm0.47$) & $58.9$ ($\pm0.66$) & $52.88$ ($\pm1.9$) & $55.72$ ($\pm1.29$)\\
300-shot & $74.44$ ($\pm2.31$) & $59.78$ ($\pm4.19$) & $56.03$ ($\pm3.58$) & $57.79$ ($\pm3.28$)\\
500-shot & $75.15$ ($\pm0.76$) & $61.1$ ($\pm1.56$) & $56.27$ ($\pm0.24$) & $58.58$ ($\pm0.81$) \\
1000-shot & $75.03$ ($\pm0.33$) & $61.01$ ($\pm0.61$) & $55.48$ ($\pm1.9$) & $58.11$ ($\pm0.74$) \\
Full-shot (1479) & $77.74$ ($\pm0.79$) & $65.17$ ($\pm2.1$) & $61.86$ ($\pm1.43$) & $63.44$ ($\pm0.34$) \\
\vspace{0.1cm}\\
\textit{Transformer - Multi-Message Input}\\
Full-shot 3 Msg & $68.76$ ($\pm0.52$) & $61.53$ ($\pm1.29$) & $59.94$ ($\pm2.02$) & $60.69$ ($\pm0.41$) \\
\textbf{Full-shot 5 Msg} & $69.77$ ($\pm0.51$) & $69.63$ ($\pm0.77$) & $76.07$ ($\pm2.65$) & $72.68$ ($\pm0.92$) \\
\hline
\end{tabular}
\caption{Performance of classification models for \textbf{Testing Boundaries}. Full-shot value refers to the number of positive examples present in the full training set. Best performing model is in bold.}
\label{tab:app-testing-boundaries-res}
\end{table*}

\begin{table*}[ht]
\centering
\begin{tabular}{lcccc}
\hline
\textbf{Model} & \textbf{Accuracy} & \textbf{Precision} & \textbf{Recall} & \textbf{F1}\\
\hline
\vspace{0.1cm}\\
\textit{Supervised - Bag of Words}\\
Random Forest & $84.63$ ($\pm0.81$) & $71.73$ ($\pm4.3$) & $26.34$ ($\pm4.31$) & $38.43$ ($\pm5.12$)\\
Logistic Regression & $85.22$ ($\pm0.73$) & $74.59$ ($\pm2.49$) & $29.17$ ($\pm5.26$) & $41.76$ ($\pm5.52$) \\
Support Vector Machine & $84.38$ ($\pm0.38$) & $81.49$ ($\pm1.64$) & $18.95$ ($\pm2.42$) & $30.7$ ($\pm3.23$)\\
Naive Bayes & $85.12$ ($\pm0.63$) & $75.04$ ($\pm4.48$) & $28.09$ ($\pm3.03$) & $40.81$ ($\pm3.45$)

\vspace{0.3cm}\\
\textit{Transformer}\\
% 0-shot & $38.01$ & $18.97$ & $72.98$ & $30.12$ \\
% 25-shot & $86.86$ & $69.44$ & $50.4$ & $58.41$ \\
% 50-shot &  $87.01$ & $70.46$ & $50.0$ & $58.49$\\
% 100-shot & $87.01$ & $67.82$ & $55.24$ & $60.89$ \\
% 200-shot & $86.5$ & $65.85$ & $54.44$ & $59.6$ \\
% 300-shot & $86.86$ & $66.2$ & $57.66$ & $61.64$\\
% 500-shot &  $85.76$ & $64.4$ & $57.58$ & $61.47$\\
% 1000-shot & $85.83$ & $63.73$ & $52.42$ & $57.52$\\
% Full-shot & $86.27$ & $64.35$ & $56.05$ & $59.91$\\

0-shot & $38.23$ ($\pm0.52$) & $19.58$ ($\pm0.73$) & $76.48$ ($\pm3.88$) & $31.18$ ($\pm1.24$)\\
25-shot & $75.57$ ($\pm2.1$) & $35.92$ ($\pm3.11$) & $41.53$ ($\pm2.02$) & $38.4$ ($\pm0.9$)\\
50-shot & $84.85$ ($\pm2.09$) & $60.61$ ($\pm7.39$) & $50.94$ ($\pm3.75$) & $55.25$ ($\pm4.59$)\\
100-shot & $85.76$ ($\pm1.02$) & $63.77$ ($\pm3.86$) & $51.62$ ($\pm1.61$) & $57.04$ ($\pm2.43$)\\
200-shot & $85.66$ ($\pm0.65$) & $63.72$ ($\pm3.64$) & $51.08$ ($\pm2.83$) & $56.58$ ($\pm0.33$)\\
300-shot & $86.0$ ($\pm0.9$) & $64.97$ ($\pm3.56$) & $51.21$ ($\pm2.82$) & $57.25$ ($\pm2.66$)\\
500-shot & $86.4$ ($\pm0.42$) & $65.17$ ($\pm1.03$) & $55.11$ ($\pm2.46$) & $59.71$ ($\pm1.78$) \\
% 1000-shot & $85.34$ ($\pm0.37$) & $61.33$ ($\pm1.33$) & $53.9$ ($\pm1.23$) & $57.37$ ($\pm0.97$)\\
1000-shot* \\
Full-shot (867) & $85.9$ ($\pm0.15$) & $61.86$ ($\pm0.33$) & $55.48$ ($\pm1.3$) & $60.88$ ($\pm0.73$) \\
\vspace{0.1cm}\\
\textit{Transformer - Multi-Message Input}\\
Full-shot 3 Msg & $84.11$ ($\pm0.51$) & $65.83$ ($\pm1.0$) & $62.59$ ($\pm1.79$) & $64.16$ ($\pm1.37$) \\
\textbf{Full-shot 5 Msg} & $79.46$ ($\pm1.08$) & $64.53$ ($\pm1.63$) & $69.26$ ($\pm3.58$) & $66.78$ ($\pm2.11$) \\
\hline
\end{tabular}
\caption{Performance of classification models for \textbf{Use of Sexual Topics}. Full-shot value refers to the number of positive examples present in the full training set. Best performing model is in bold.}
{\raggedright * The number exceeds the total number of positive examples in the dataset and therefore results are analogous to full-shot \par}
\label{tab:app-using-sex-res}
\end{table*}

\begin{table*}[ht]
\centering
\begin{tabular}{lcccc}
\hline
\textbf{Model} & \textbf{Accuracy} & \textbf{Precision} & \textbf{Recall} & \textbf{F1}\\
\hline
\vspace{0.1cm}\\
\textit{Supervised - Bag of Words}\\
Random Forest & $96.97$ ($\pm0.2$) & $44.44$ ($\pm38.49$) & $4.88$ ($\pm4.88$) & $8.7$ ($\pm8.5$)\\
Logistic Regression & $96.93$ ($\pm0.19$) & $42.05$ ($\pm20.31$) & $10.57$ ($\pm7.04$) & $16.7$ ($\pm10.72$) \\
Support Vector Machine & $96.97$ ($\pm0.01$) & $0.0$ ($\pm0.0$) & $0.0$ ($\pm0.0$) & $0.0$ ($\pm0.0$)\\
Naive Bayes & $97.02$ ($\pm0.04$) & $58.89$ ($\pm8.39$) & $4.87$ ($\pm2.44$) & $8.93$ ($\pm4.2$)

\vspace{0.3cm}\\
\textit{Transformer}\\
% 0-shot &  $44.8$ & $3.3$ & $60.98$ & $6.27$ \\
% 25-shot & $96.68$ & $30.0$ & $7.32$ & $11.65$ \\
% 50-shot &  $96.61$ & $40.0$ & $24.39$ & $30.3$\\
% 100-shot & $96.75$ & $41.18$ & $17.07$ & $24.14$ \\
% 200-shot & $97.05$ & $52.94$ & $21.95$ & $31.03$\\
% 300-shot* & \\
% 500-shot* &  \\
% 1000-shot* & \\
% Full-shot & $97.27$ & $57.14$ & $39.02$ & $46.38$ \\

0-shot & $44.11$ ($\pm0.89$) & $3.22$ ($\pm0.23$) & $60.16$ ($\pm0.43$) & $6.12$ ($\pm0.43$)\\
25-shot & $96.73$ ($\pm0.33$) & $40.3$ ($\pm12.74$) & $13.01$ ($\pm2.82$) & $19.38$ ($\pm3.86$)\\
50-shot & $97.34$ ($\pm0.39$) & $66.52$ ($\pm17.64$) & $23.58$ ($\pm8.57$) & $34.7$ ($\pm11.48$)\\
100-shot & $97.32$ ($\pm0.01$) & $64.08$ ($\pm2.95$) & $26.83$ ($\pm7.32$) & $37.28$ ($\pm6.8$)\\
200-shot* \\
300-shot* & \\
500-shot* &  \\
1000-shot* & \\
Full-shot (144) & $96.78$ ($\pm0.15$) & $44.9$ ($\pm4.34$) & $30.08$ ($\pm5.08$) & $36.0$ ($\pm5.06$)\\
\vspace{0.1cm}\\
\textit{Transformer - Multi-Message Input}\\
\textbf{Full-shot 3 Msg} & $96.75$ ($\pm0.51$) & $62.22$ ($\pm10.72$) & $38.39$ ($\pm7.0$) & $47.47$ ($\pm8.45$)\\
Full-shot 5 Msg & $94.83$ ($\pm0.11$) & $50.88$ ($\pm1.52$) & $38.27$ ($\pm2.14$) & $43.64$ ($\pm1.18$)\\
\hline
\end{tabular}
\caption{Performance of classification models for \textbf{Mitigation/Minimization}. Full-shot value refers to the number of positive examples present in the full training set. Best performing model is in bold.}
{\raggedright * The number exceeds the total number of positive examples in the dataset and therefore results are analogous to full-shot \par}
\label{tab:app-mitigate-res}
\end{table*}

% Dynamic Identities
%  Removed for the moment as I'm unsure how it can be described in this work. Might remove.

\begin{table*}[ht]
\centering
\begin{tabular}{lcccc}
\hline
\textbf{Model} & \textbf{Accuracy} & \textbf{Precision} & \textbf{Recall} & \textbf{F1}\\
\hline
\vspace{0.1cm}\\
\textit{Supervised - Bag of Words}\\
Random Forest & $91.51$ ($\pm0.27$) & $36.6$ ($\pm6.36$) & $7.03$ ($\pm1.91$) & $11.67$ ($\pm2.59$)\\
Logistic Regression & $91.61$ ($\pm0.3$) & $30.51$ ($\pm15.39$) & $3.36$ ($\pm1.91$) & $6.05$ ($\pm3.39$) \\
Support Vector Machine & $91.88$ ($\pm0.01$) & $42.11$ ($\pm11.07$) & $3.67$ ($\pm2.43$) & $6.67$ ($\pm4.32$) \\
Naive Bayes & $91.88$ ($\pm0.07$) & $27.78$ ($\pm25.46$) & $0.61$ ($\pm0.53$) & $1.2$ ($\pm1.0$)

\vspace{0.3cm}\\
\textit{Transformer}\\
% 0-shot & $61.18$ & $11.46$ & $56.88$ & $19.08$ \\
% 25-shot & $90.48$ & $30.77$ & $14.68$ & $19.88$ \\
% 50-shot & $89.08$ & $25.32$ & $18.35$ & $21.28$ \\
% 100-shot & $89.89$ & $30.0$ & $19.27$ & $23.46$ \\
% 200-shot & $90.26$ & $31.75$ & $18.35$ & $23.26$\\
% 300-shot & $90.48$ & $32.76$ & $17.43$ & $22.76$\\
% 500-shot & $90.85$ & $38.81$ & $23.85$ & $29.55$ \\
% 1000-shot* & \\
% Full-shot & $90.7$ & $38.67$ & $26.61$ & $31.52$ \\

0-shot & $61.62$ ($\pm0.9$) & $11.49$ ($\pm1.26$) & $56.27$ ($\pm6.44$) & $19.08$ ($\pm2.1$)\\
25-shot & $88.98$ ($\pm0.37$) & $18.36$ ($\pm3.51$) & $10.7$ ($\pm1.91$) & $13.52$ ($\pm2.48$)\\
50-shot & $90.65$ ($\pm0.87$) & $37.07$ ($\pm7.77$) & $21.1$ ($\pm0.92$) & $26.72$ ($\pm2.11$)\\
100-shot & $89.91$ ($\pm0.6$) & $30.33$ ($\pm4.72$) & $19.88$ ($\pm1.84$) & $23.55$ ($\pm2.79$)\\
200-shot & $91.61$ ($\pm0.04$) & $45.13$ ($\pm0.7$) & $19.88$ ($\pm1.4$) & $27.58$ ($\pm1.44$)\\
\textbf{300-shot} & $91.37$ ($\pm0.46$) & $43.16$ ($\pm4.83$) & $22.02$ ($\pm1.59$) & $29.1$ ($\pm2.04$)\\
500-shot* & \\
1000-shot* & \\
Full-shot (380) & $91.14$ ($\pm0.22$) & $39.88$ ($\pm2.78$) & $19.88$ ($\pm1.4$) & $26.53$ ($\pm1.89$)\\

\vspace{0.1cm}\\
\textit{Transformer - Multi-Message Input}\\
Full-shot 3 Msg & $86.43$ ($\pm0.2$) & $33.99$ ($\pm1.09$) & $18.0$ ($\pm0.01$) & $23.53$ ($\pm0.27$)\\
Full-shot 5 Msg & $78.75$ ($\pm1.4$) & $33.08$ ($\pm4.8$) & $22.47$ ($\pm3.89$) & $26.69$ ($\pm4.02$)\\
\hline
\end{tabular}
\caption{Performance of classification models for \textbf{Encouragement}. Full-shot value refers to the number of positive examples present in the full training set. Best performing model is in bold.}
{\raggedright * The number exceeds the total number of positive examples in the dataset and therefore results are analogous to full-shot \par}
\label{tab:app-encourage-res}
\end{table*}

\begin{table*}[ht]
\centering
\begin{tabular}{lcccc}
\hline
\textbf{Model} & \textbf{Accuracy} & \textbf{Precision} & \textbf{Recall} & \textbf{F1}\\
\hline
\vspace{0.1cm}\\
\textit{Supervised - Bag of Words}\\
Random Forest & $95.75$ ($\pm0.35$) & $63.83$ ($\pm17.9$) & $22.04$ ($\pm6.52$) & $31.79$ ($\pm6.45$)\\
Logistic Regression & $95.7$ ($\pm0.3$) & $62.86$ ($\pm14.5$) & $18.28$ ($\pm4.06$) & $27.81$ ($\pm3.99$) \\
Support Vector Machine & $95.47$ ($\pm0.01$) & $54.37$ ($\pm25.54$) & $4.3$ ($\pm3.36$) & $7.89$ ($\pm5.93$)\\
Naive Bayes & $95.57$ ($\pm0.13$) & $67.14$ ($\pm15.45$) & $6.45$ ($\pm1.61$) & $11.75$ ($\pm2.84$)

\vspace{0.3cm}\\
\textit{Transformer}\\
0-shot & $48.61$ ($\pm0.52$) & $4.75$ ($\pm0.35$) & $53.76$ ($\pm4.06$) & $8.74$ ($\pm0.65$)\\
25-shot & $95.25$ ($\pm0.09$) & $44.74$ ($\pm2.35$) & $15.59$ ($\pm1.86$) & $23.07$ ($\pm10.94$)\\
50-shot & $95.84$ ($\pm0.57$) & $56.11$ ($\pm9.48$) & $38.17$ ($\pm10.74$) & $45.27$ ($\pm10.65$)\\
100-shot & $96.11$ ($\pm0.37$) & $63.04$ ($\pm8.13$) & $37.1$ ($\pm6.45$) & $46.49$ ($\pm6.06$)\\
\textbf{200-shot} & $96.36$ ($\pm0.26$) & $63.08$ ($\pm3.97$) & $49.46$ ($\pm4.06$) & $55.39$ ($\pm3.48$)\\
300-shot* &\\
500-shot* &  \\
1000-shot* & \\
Full-shot (218) & $96.29$ ($\pm0.09$) & $65.85$ ($\pm2.39$) & $39.25$ ($\pm0.93$) & $49.16$ ($\pm0.58$)\\
\vspace{0.1cm}\\
\textit{Transformer - Multi-Message Input}\\
Full-shot 3 Msg & $94.47$ ($\pm0.44$) & $61.83$ ($\pm6.42$) & $29.63$ ($\pm6.42$) & $39.98$ ($\pm7.28$)\\
Full-shot 5 Msg & $92.64$ ($\pm0.39$) & $65.48$ ($\pm3.21$) & $47.62$ ($\pm1.18$) & $55.13$ ($\pm1.86$)\\
\hline
\end{tabular}
\caption{Performance of classification models for \textbf{Risk Management}. Full-shot value refers to the number of positive examples present in the full training set. Best performing model is in bold.}
{\raggedright * The number exceeds the total number of positive examples in the dataset and therefore results are analogous to full-shot \par}
\label{tab:app-risk-management}
\end{table*}

\end{document}